\documentclass[english]{article}
\usepackage[utf8]{inputenc}
\usepackage[T1]{fontenc}
\usepackage{babel}
\usepackage{amsmath}
\usepackage{graphicx}
\usepackage{rotating}
\usepackage{fancyhdr}
\usepackage[hyphens]{url}

\begin{document}

\title{A large annotated medical image dataset for the development and evaluation of segmentation algorithms}

\author{\parbox{\linewidth}{\centering
        Amber L. Simpson\textsuperscript{1{*}}, 
        Michela Antonelli\textsuperscript{2}, 
        Spyridon Bakas\textsuperscript{3}, 
        Michel Bilello\textsuperscript{3}, 
        Keyvan Farahani\textsuperscript{4}, 
        Bram van Ginneken\textsuperscript{5}, 
        Annette Kopp-Schneider\textsuperscript{6}, 
        Bennett A. Landman\textsuperscript{7}, 
        Geert Litjens\textsuperscript{5}, 
        Bjoern Menze\textsuperscript{8}, 
        Olaf Ronneberger\textsuperscript{9}, 
        Ronald M. Summers\textsuperscript{10}, 
        Patrick Bilic\textsuperscript{8}, 
        Patrick F. Christ\textsuperscript{8}, 
        Richard K. G. Do\textsuperscript{11}, 
        Marc Gollub\textsuperscript{11}, 
        Jennifer Golia-Pernicka\textsuperscript{11}, 
        Stephan H. Heckers\textsuperscript{12}, 
        William R. Jarnagin\textsuperscript{1}, 
        Maureen K. McHugo\textsuperscript{12}, 
        Sandy Napel\textsuperscript{13}, 
        Eugene Vorontsov\textsuperscript{14}, 
        Lena Maier-Hein\textsuperscript{15}, and
        M. Jorge Cardoso\textsuperscript{16}
} % end centering
} % end author

\maketitle
\thispagestyle{fancy}

1. Department of Surgery, Memorial Sloan Kettering Cancer Center.
2. Centre for Medical Image Computing, University College London.
3. Center for Biomedical Image Computing and Analytics, University of Pennsylvania.
4. Division of Cancer Treatment and Diagnosis, National Cancer Institute.
5. Department of Pathology, Radboud University Medical Center.
6. Division of Biostatistics, German Cancer Research Center.
7. Department of Electrical Engineering and Computer Science, Vanderbilt University.
8. Department of Informatics, Technische Universität München. 
9. Google DeepMind. 
10. Imaging Biomarkers and Computer-aided Diagnosis Lab, Radiology and Imaging Sciences, National Institutes of Health Clinical Center.
11. Department of Radiology, Memorial Sloan Kettering Cancer Center.
12. Department of Psychiatry \& Behavioral Sciences, Vanderbilt University Medical Center.
13. Department of Radiology, Stanford University.
14. Department of Computer Science and Software Engineering, École Polytechnique de Montréal.
15. Division of Computer Assisted Medical Interventions, German Cancer Research Center.
16. Department of Imaging and Biomedical Engineering, King's College London.
{*}corresponding author:
Amber Simpson (simpsonl@mskcc.org)

\begin{abstract}
Semantic segmentation of medical images aims to associate a pixel with a label in a medical image without human initialization. The success of semantic segmentation algorithms is contingent on the availability of high-quality imaging data with corresponding labels provided by experts. We sought to create a large collection of annotated medical image datasets of various clinically relevant anatomies available under open source license to facilitate the development of semantic segmentation algorithms. Such a resource would allow: 1) objective assessment of general-purpose segmentation methods through comprehensive benchmarking and 2) open and free access to medical image data for any researcher interested in the problem domain. Through a multi-institutional effort, we generated a large, curated dataset representative of several highly variable segmentation tasks that was used in a crowd-sourced challenge – the Medical Segmentation Decathlon held during the 2018 Medical Image Computing and Computer Aided Interventions Conference in Granada, Spain. Here, we describe these ten labeled image datasets so that these data may be effectively reused by the research community.
\end{abstract}

\section*{Background \& Summary}

Medical image segmentation seeks to extract, either semi-automatically or automatically, anatomical regions of interest from a medical image or series of images. Specific regions of interest can range from tumours to bone to blood vessels, depending on the clinical application. Medical image segmentation algorithms have been proposed for decades; however, almost none have been integrated into clinical systems. Consequently, clinicians are routinely required to demarcate regions of interest manually for a variety of clinical applications, including treatment planning or volumetric measurements of a tumour to assess response to therapy. Semantic segmentation aims to automatically associate a pixel with a label in an image without any initialization~\cite{Shelhamer2017}. Semantic segmentation algorithms are becoming increasingly general purpose and translatable to unseen tasks~\cite{Guo2018}. A fully automated semantic segmentation model that works out-of-the-box on many tasks, in the spirit of automated machine learning (AutoML), would have a tremendous impact on healthcare. 

With sufficient medical imaging data representative of a given task, a general purpose semantic segmentation algorithm is potentially extendable to new tasks, overcoming traditional limitations due to anatomical variation across different patients~\cite{Roth2018}. The success of semantic segmentation of medical images is contingent on the availability of high-quality labeled medical image data. Institutions are reluctant to share imaging data for a variety of reasons~\cite{Chatterjee2017}, most notably for privacy concerns. Strict health information privacy regulations require the removal of protected health information (PHI) before sharing data outside of the home institution, a process that in the context of medical image data can be expensive and time-consuming. Institutions that release PHI, either knowingly or unknowingly, are potentially subject to serious consequences including substantial fines and criminal prosecution. In many cases, when imaging data are publicly accessible (e.g., The Cancer Imaging Archive~\cite{Prior2017}), the corresponding labels are rarely available due to the lack of expert annotations of the regions of interest and a lack of infrastructure and standards for sharing labeled data.

Recent advances in machine learning have led to a sharp increase in the number of medical image segmentation algorithms reported in the literature. Many key algorithmic enhancements in the field are validated on a relatively small number of samples~\cite{Maier-Hein2018} with only a few comparisons to existing approaches, thereby limiting our understanding of the generalizability of the proposed algorithms and our ability to differentiate novel from incremental advances. In the computer vision field, the PASCAL Visual Object Classes (VOC) project largely solved the object recognition problem by providing a publicly available labeled dataset for benchmarking through crowdsourcing~\cite{Everingham2014}. Inspired by the progress achieved by the PASCAL VOC project, we sought to create a large, open source, manually annotated medical image dataset of various anatomical sites that would enable objective assessment of general-purpose segmentation methods through comprehensive benchmarking and would democratize access to medical image data. We embarked on a multi-institutional collaboration to generate a sufficiently large dataset consisting of several highly variable segmentation tasks. These data were used in a crowd-sourced challenge of generalizable semantic segmentation algorithms called the Medical Segmentation Decathlon (MSD) held during the 2018 Medical Image Computing and Computer Aided Interventions (MICCAI) Conference in Granada, Spain. This paper details the ten labeled medical image datasets used in the challenge and their availability to the research community.

\section*{Methods}

In total, 2,633 three-dimensional images were collected across multiple anatomies of interest, multiple modalities, and multiple sources (or institutions) representative of real-world clinical applications. All images were de-identified using processes consistent with institutional review board polices at each contributing site. We reformatted the images to reduce the need for specialized software packages for reading to encourage use by the general machine learning community, not only specialists in medical imaging. 

\subsection*{Datasets}

\begin{figure}[ht]
\includegraphics[width=\textwidth]{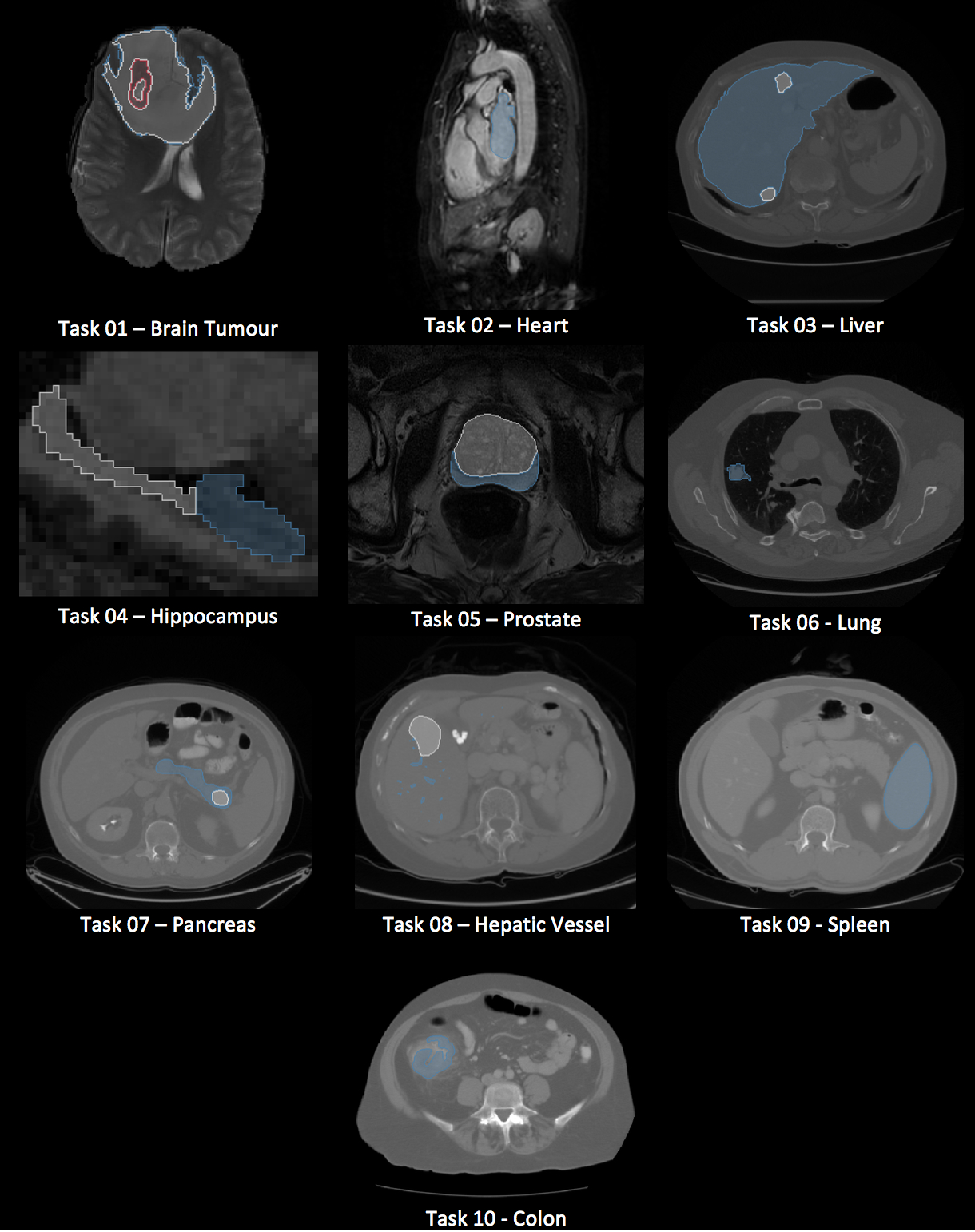}
\caption{Exemplar images and labels for each dataset. Blue, white, and red correspond to labels 1, 2, and 3, respectively, of each dataset. Not all tasks have 3 labels.}~\label{fig:data}
\end{figure}

An overview of the ten datasets is provided in Table~\ref{tab:data}. The datasets were chosen based on availability and appropriateness for semantic segmentation algorithm development. The rationale for inclusion of each dataset is described in the table. Exemplar images and labels for each task are provided in Figure~\ref{fig:data}.

\paragraph*{Task01\_BrainTumour} The brain tumour dataset included in the MSD challenge describes a subset of the data used in the 2016 and 2017 Brain Tumour Image Segmentation (BraTS) challenges~\cite{Bakas2018,Bakas2017,Menze2015}. Specifically, 750 multi-parametric magnetic resonance imaging (MRI) scans from patients diagnosed with either glioblastoma or lower-grade glioma were included. The multi-para\-metric MRI sequences of each patient included native (T1) and post-Gadolinium (Gd) contrast T1-weighted (T1-Gd), native T2-weighted (T2), and T2 Fluid-Attenuated Inversion Recovery (T2-FLAIR) volumes. These MRI scans were acquired during routine clinical practice, using different equipment and acquisition protocols, among 19 different institutions and pooled to create a publicly available benchmark dataset for the task of segmenting brain tumour sub-regions (i.e., edema, enhancing, and non-enhancing tumour). The scanners used for the acquisition of these scans varied from 1T to 3T. All scans were co-registered to a reference atlas space using the SRI24 brain structure template~\cite{Bakas2018}, resampled to isotropic voxel resolution of 1 mm$^3$, and skull-stripped using various methods followed by manual refinements. Gold standard annotations for all tumour sub-regions in all scans were approved by expert board-certified neuroradiologists. 

Data contributing institutions for the BraTS dataset included: 1) Center for Biomedical Image Computing and Analytics (CBICA), University of Pennsylvania, PA, USA, 2) University of Alabama at Birmingham, AL, USA, 3) Heidelberg University, Germany, 4) University Hospital of Bern, Switzerland, 5) University of Debrecen, Hungary, 6) Henry Ford Hospital, MI, USA, 7) University of California, CA, USA, 8) MD Anderson Cancer Center, TX, USA, 9) Emory University, GA, USA, 10) Mayo Clinic, MN, USA, 11) Thomas Jefferson University, PA, USA, 12) Duke University School of Medicine, NC, USA, 13) Saint Joseph Hospital and Medical Center, AZ, USA, 14) Case Western Reserve University, OH, USA, 15) University of North Carolina, NC, USA, 16) Fondazione IRCCS Istituto Neurologico Carlo Besta, Italy, 17) Washington University School of Medicine in St. Louis, MO, USA, and 18) Tata Memorial Centre, Mumbai, India. Data from institutions 6-16 describe data from TCIA (http://www.cancerimagingarchive.net/)~\cite{Clark2013}.

\paragraph*{Task02\_Heart} The heart dataset was provided by King’s College London (London, United Kingdom), originally released through the Left Atrial Segmentation Challenge (LASC)~\cite{Tobon-Gomez2015} and includes 30 MRI datasets covering the entire heart acquired during a single cardiac phase (free breathing with respiratory and ECG gating). Images were obtained on a 1.5T Achieva scanner (Philips Healthcare, Best, The Netherlands) with voxel resolution 1.25 x 1.25 x 2.7 mm$^3$. The left atrium appendage, mitral plane, and portal vein end points were segmented by an expert using an automated tool~\cite{Ecabert2011} followed by manual correction. 

\paragraph*{Task03\_Liver} This liver dataset consisted of 201 contrast-enhanced CT images originally provided from several clinical sites, including Ludwig Maximilian University of Munich (Germany), Radboud University Medical Center of Nijmegen (The Netherlands), Polytechnique and CHUM Research Center Montreal (Canada), Tel Aviv University (Israel), Sheba Medical Center (Israel), IRCAD Institute Strasbourg (France), and Hebrew University of Jerusalem (Israel) through the Liver Tumour Segmentation (LiTS) challenge~\cite{Bilic2019}. The patients included had a variety of primary cancers, including hepatocellular carcinoma, as well as metastatic liver disease derived from colorectal, breast, and lung primary cancers. CT scans included a variety of pre- and post-therapy images. Some images contained metal artifacts, consistent with real-world clinical scenarios for abdominal CT. The images were provided with an in-plane resolution of 0.5 to 1.0 mm, and slice thickness of 0.45 to 6.0 mm. Annotations of the liver and tumours were performed by radiologists. 

\paragraph*{Task04\_Hippocampus} The dataset consisted of MRI acquired in 90 healthy adults and 105 adults with a non-affective psychotic disorder (56 schizophrenia, 32 schizoaffective disorder, and 17 schizophreniform disorder) taken from the Psychiatric Genotype/Phenotype Project data repository at Vanderbilt University Medical Center (Nashville, TN, USA). Patients were recruited from the Vanderbilt Psychotic Disorders Program and controls were recruited from the surrounding community. All participants were assessed with the Structured Clinical Interview for DSM-IV~\cite{First2002}. All subjects were free from significant medical or neurological illness, head injury, and active substance use or dependence. 

Structural images were acquired with a 3D T1-weighted MPRAGE sequence (TI/TR/TE, 860/8.0/3.7 ms; 170 sagittal slices; voxel size, 1.0 mm$^3$). All images were collected on a Philips Achieva scanner (Philips Healthcare, Inc., Best, The Netherlands). Manual tracing of the head, body, and tail of the hippocampus on images was completed following a previously published protocol~\cite{Pruessner2000,Woolard2012}. For the purposes of this dataset, the term hippocampus includes the hippocampus proper (CA1-4 and dentate gyrus) and parts of the subiculum, which together are more often termed the hippocampal formation~\cite{Amaral1989}. The last slice of the head of the hippocampus was defined as the coronal slice containing the uncal apex. The resulting 195 labeled images are referred to as hippocampus atlases. Note that the term hippocampus posterior refers to the union of the body and the tail. 

\paragraph*{Task05\_Prostate} The prostate dataset consisted of 48 multi-para\-metric MRI studies provided by Radboud University (The Netherlands) reported in a previous segmentation study~\cite{Litjens2012}. Manual segmentation of the whole prostate from transverse T2-weighted scans with resolution 0.6 x 0.6 x 4 mm and the apparent diffusion coefficient (ADC) map (2 x 2 x 4 mm) was used. 

\paragraph*{Task06\_Lung} The lung dataset was comprised of patients with non-small cell lung cancer from Stanford University (Palo Alto, CA, USA) publicly available through TCIA and previously utilized to create a radiogenomic signature~\cite{Napel2014,Bakr2018,Gevaert2012}. Briefly, 96 preoperative thin-section CT scans were obtained with the following acquisition and reconstruction parameters: section thickness, <1.5 mm; 120 kVp; automatic tube current modulation range, 100–700 mA; tube rotation speed, 0.5 s; helical pitch, 0.9-1.0; and a sharp reconstruction kernel. The tumour region was denoted by an expert thoracic radiologist on a representative CT cross section using OsiriX~\cite{Rosset2004}.

\paragraph*{Task07\_Pancreas} The pancreas dataset was comprised of patients undergoing resection of pancreatic masses (intraductal mucinous neoplasms, pancreatic neuroendocrine tumours, or pancreatic ductal adenocarcinoma). Images were provided by Memorial Sloan Kettering Cancer Center (New York, NY, USA) and were previously reported in radiomic applications~\cite{Attiyeh2018,Attiyeh2018a,Chakraborty2018}. Four hundred and twenty portal venous phase CT scans were obtained with the following reconstruction and acquisition parameters:  pitch/table speed 0.984–1.375/39.37–27.50 mm; automatic tube current modulation range, 220–380 mA; noise index, 12.5–14; 120 kVp; tube rotation speed, 0.7–0.8 ms; scan delay, 80–85 s; and axial slices reconstructed at 2.5 mm intervals. The pancreatic parenchyma and pancreatic mass (cyst or tumour) were manually segmented in each slice by an expert abdominal radiologist using the Scout application~\cite{Dawant2007}.

\paragraph*{Task08\_HepaticVessel} This second liver dataset was comprised of patients with a variety of primary and metastatic liver tumours (cholangiocarcinoma, hepatocellular carcinoma, metastases, etc.) provided by Memorial Sloan Kettering Cancer Center (New York, NY, USA) and previously reported~\cite{Pak2018,Simpson2015,Simpson2017,Zheng2017}. Four hundred and forty-three portal venous phase CT scans were obtained with the following criteria: 120 kVp; exposure time, 500–1100 ms; and tube current, 33–440 mA. Images were reconstructed at a section thickness varying from 2.5 to 5 mm with a standard convolutional kernel and with a reconstruction diameter range of 360–500 mm. Iodinated contrast material (150 mL, Omnipaque 300, GE Healthcare, Chicago, IL, USA) was administered intravenously for each CT at a rate between 1 and 4 cc/s. The liver vessels were semi-automatically segmented using the Scout application~\cite{Dawant2007}. Briefly, a seed point was drawn on the region of interest and grown using a level-set based approach. Contours were manually adjusted by an expert abdominal radiologist.  

\paragraph*{Task09\_Spleen} The spleen dataset was comprised of patients undergoing chemotherapy treatment for liver metastases at Memorial Sloan Kettering Cancer Center (New York, NY, USA) and previously reported~\cite{Simpson2015spleen}. Sixty-one portal venous phase CT scans were included with CT acquisition and reconstruction parameters similar to the Task08\_HepaticVessel dataset above. The spleen was semi-automatically segmented using the Scout application~\cite{Dawant2007}. A spline was drawn on the region of interest and grown using a level-set based approach. Contours were manually adjusted by an expert abdominal radiologist.  

\paragraph{Task10\_Colon} The colon dataset was comprised of patients undergoing resection of primary colon cancer at Memorial Sloan Kettering Cancer Center (New York, NY, USA). Many of the scans were ordered by a referring physician at an outside institution and transferred to MSK as part of routine clinical care for colon cancer patients referred to the center. Consequently, the scan protocols varied widely. One hundred and ninety portal venous phase CT scans were obtained with different scanner manufacturers and different contrast agents. Acquisition parameters were: 100–140 kVp; exposure time, 500–1782 ms; and tube current, 100–752 mA.  Reconstruction parameters were: slice thickness, 1 to 7.5 mm; and reconstruction diameter, 274–500 mm.  The colon tumour was manually segmented using ITK Snap (Philadelphia, PA) by an expert radiologist in body imaging. 

\subsection*{Data Processing}

Data were reformatted to ensure ease of access by enforcing consistency and interoperability across all datasets. We chose to reformat all images from standard DICOM to Neuroimaging Informatics Technology Initiative (NIfTI) images, an open format supported by the National Institutes of Health~\cite{nifti}. Therefore, no proprietary software is needed to utilize the data. 

\subsection*{Data Availability}

All data are downloadable from \url{http://medicaldecathlon.com/}.

\subsection*{Data Licensing}

All data were made available online under Creative Commons license CC-BY-SA 4.0, allowing the data to be shared or redistributed in any format and improved upon, with no commercial restrictions~\cite{Chatterjee2017}. Under this license, the appropriate credit must be given (by citation to this paper), with a link to the license and any changes noted. The images can be redistributed under the same license.
 
\subsection*{Data Records}

\begin{figure}[ht]
\includegraphics[width=\textwidth]{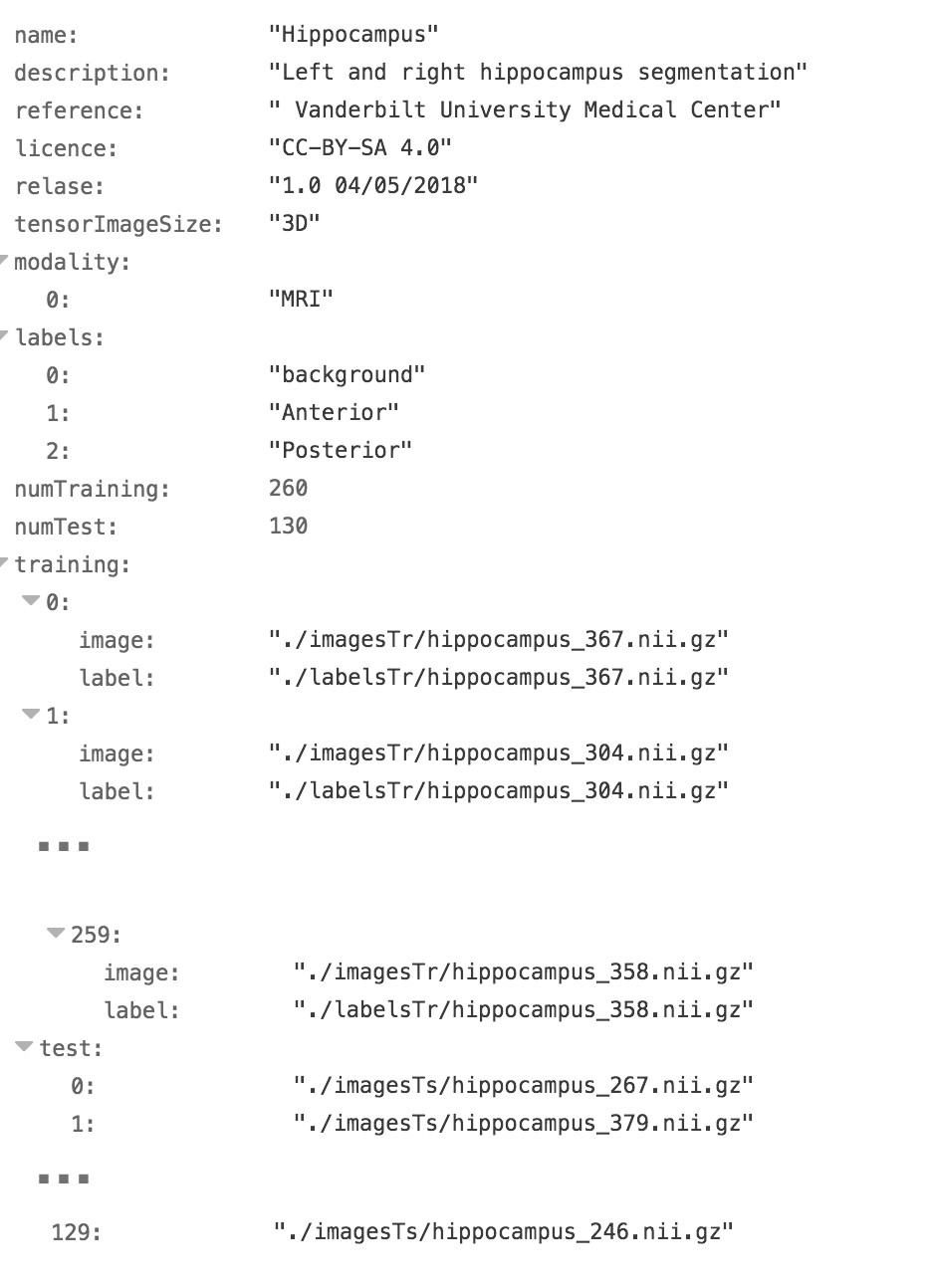}
\caption{Example JSON file included in the download for each dataset. }~\label{fig:JSON}
\end{figure}

Each of the ten datasets is downloadable as a TAR file that includes a JSON descriptor as well as directories containing the images and labels for the training set for the MSD challenge, and the images in the testing set (all images in NIfTI format). The JSON file includes information necessary to use the dataset, including the imaging modality, the number of training and testing images, and the names of all images and corresponding labels. The JSON file for the hippocampus dataset is provided in Figure~\ref{fig:JSON} as an example. Each image is labeled systematically according to the dataset name followed by a unique identifier. For example, ‘hippocampus\_367.nii’ is Subject 367’s image in the hippocampus dataset; the image (from the training set) is stored in the imagesTr directory, and the label is stored in the labelsTr directory.

\subsection*{Technical Validation}

All imaging data were acquired through routine clinical scanning and analyzed retrospectively. Therefore, there is substantial variation in acquisition and reconstruction parameters due to the variability in protocols within and across institutions that represents real-world clinical imaging scenarios. The technical quality of the individual datasets is evidenced by the prior works reporting use as referenced in Table~\ref{tab:data}.  

The images were reformatted to NIfTI into a standard anatomical coordinate system, potentially introducing coordinate transformation errors due to inconsistency in the DICOM coordinate frame. Each image was manually verified by the second author to rectify these errors. All images were reformatted to a Right-Anterior-Superior (RAS) coordinate frame, allowing algorithms to assume that the xyz millimeter coordinates have the same order and direction as the voxels ijk coordinates. Some images in the Task 1 and Task 2 datasets required a further rotation by 180 degrees within the axial plan to correct for misformatted headers, matching the orientation of the remaining images in the task. All non-quantitative magnitude images were linearly scaled using a robust min-max to the 0–1000 range to avoid problems with misformatted DICOM headers missing the appropriate scaling factor; all other images (e.g., CT in Hounsfield units) remained unscaled to preserve their quantitative nature.

\subsection*{Usage Notes}

For more detailed information on NIfTI images, refer to the NIfTI website (\url{https://nifti.nimh.nih.gov/)}. Multiple software platforms can read and manipulate NIfTI images, including 3D Slicer (\url{https://www.slicer.org/}), ITK Snap (\url{http://www.itksnap.org/}), and MATLAB \sloppy{(\url{https://www.mathworks.com/help/images/ref/niftiread.html})}. Numerous studies report the use of these individual datasets for various problems ranging from prognostic model building to image segmentation (see Table~\ref{tab:data}); however, the MSD challenge was the first general-purpose segmentation challenge that attempted to incorporate all ten datasets. Multiple scanners from multiple institutions were used to acquire these data. Researchers should be mindful that these images were rotated to a standard anatomical coordinate system as described above.

\section*{Acknowledgements}

This work was supported in part by the: 1) National Institute of Neurological Disorders and Stroke (NINDS) of the NIH R01 grant with award number R01-NS042645, 2) Informatics Technology for Cancer Research (ITCR) program of the NCI/NIH U24 grant with award number U24-CA189523, 3) National Cancer Institute Cancer Center Support Grant P30 CA008748, 4) American Association of Cancer Research, and 5) Intramural Research Program of the National Institutes of Health Clinical Center (1Z01 CL040004).

\section*{Competing financial interests}

G.L. received research funding from Philips Digital Pathology Solutions (Best, the Netherlands) and has a consultancy role for Novartis (Basel, Switzerland). He received research grants from the Dutch Cancer Society (KUN 2015-7970), from Netherlands Organization for Scientific Research (NWO) (project number 016.186.152), and from Stichting IT Projecten (project PATHOLOGIE 2).

R.M.S. receives royalties from iCAD, Philips, ScanMed and PingAn, and research support from PingAn and NVIDIA.

%\section*{Figures}

%\section*{Tables}

\begin{sidewaystable}[ht]
\centering 
\begin{tabular}{l l l p{2cm} l p{2cm} l}
ID	& Target &	Modality &	Image Series &	N &	Labels	& Ref \\
\hline \hline
Task01\_BrainTumour	& Brain tumours & MRI &	FLAIR, T1w, T1gd,T2w &	750 &	Glioma (necrotic/active tumour), edema &  ~\cite{Bakas2017,Bakas2018,Menze2015} \\
\hline
Task02\_Heart &	Cardiac	& MRI & 3D & 30 &	Left atrium	& ~\cite{Ecabert2011} \\
\hline
Task03\_Liver &	Liver and tumours &	CT & Portal venous phase & 201 &	Liver, tumours	&	~\cite{Bilic2019} \\
\hline
Task04\_Hippocampus	& Hippocampus & MRI	& & 394 & Hippocampus head and body
	&	 \\	
\hline
Task05\_Prostate & Prostate	& MRI &	T2, ADC	& 48 & Prostate central gland, peripheral zone 	&	~\cite{Litjens2012} \\
\hline
Task06\_Lung & Lung tumours & CT &	Non-contrast &	96 & Lung tumours &	~\cite{Bakr2018,Gevaert2012,Napel2014} \\
\hline
Task07\_Pancreas & Pancreas	& CT & Portal venous phase & 420 & Pancreas, pancreatic mass (cyst or tumour) & ~\cite{Attiyeh2018,Attiyeh2018a,Chakraborty2018} \\
\hline
Task08\_HepaticVessel &	Liver tumours and vessels &	CT & Portal venous phase & 443 & Liver vessels	&	~\cite{Pak2018,Simpson2015,Simpson2017,Zheng2017} \\
\hline
Task09\_Spleen & Spleen & CT & Portal venous phase & 61 & Spleen & ~\cite{Simpson2015spleen} 	\\
\hline
Task10\_Colon &	Colon tumours &	CT & Non-contrast &	190	& Colon cancer \\
\hline

    \end{tabular}
    \caption{Medical image datasets available for download and reuse in this collection.}
    \label{tab:data}
\end{sidewaystable}

\bibliographystyle{naturemag}
\bibliography{references.bib}

\section*{Data Citations}

\noindent 1. Scarpace, L., Mikkelsen, T., Cha, soonmee, Rao, S., Tekchandani, S., Gutman, D., … Pierce, L. J. (2016). Radiology Data from The Cancer Genome Atlas Glioblastoma Multiforme [TCGA-GBM] collection. \emph{The Cancer Imaging Archive}. http://doi.org/10.7937/K9/TCIA.2016.RNYFUYE9 \\

\noindent 2. Pedano, N., Flanders, A. E., Scarpace, L., Mikkelsen, T., Eschbacher, J. M., Hermes, B., … Ostrom, Q. (2016). Radiology Data from The Cancer Genome Atlas Low Grade Glioma [TCGA-LGG] collection. \emph{The Cancer Imaging Archive}. http://doi.org/10.7937/K9/TCIA.2016.L4LTD3TK \\

\noindent 3. Spyridon Bakas, Hamed Akbari, Aristeidis Sotiras, Michel Bilello, Martin Rozycki, Justin Kirby, John Freymann, Keyvan Farahani, and Christos Davatzikos. (2017) Segmentation Labels and Radiomic Features for the Pre-operative Scans of the TCGA-GBM collection. \emph{The Cancer Imaging Archive}. https://doi.org/10.7937/K9/TCIA.2017.KLXWJJ1Q \\

\noindent 4. Spyridon Bakas, Hamed Akbari, Aristeidis Sotiras, Michel Bilello, Martin Rozycki, Justin Kirby, John Freymann, Keyvan Farahani, and Christos Davatzikos. (2017) Segmentation Labels and Radiomic Features for the Pre-operative Scans of the TCGA-LGG collection. \emph{The Cancer Imaging Archive}. https://doi.org/10.7937/K9/TCIA.2017.GJQ7R0EF \\

\end{document}